\pdfoutput=1

\documentclass[11pt]{article}
\usepackage{multirow}
\usepackage{multicol}
\usepackage{ACL2023}
\usepackage{graphicx}
\usepackage{times}
\usepackage{latexsym}
\usepackage{url}
\usepackage[T1]{fontenc}
\usepackage{amsmath}
\usepackage{amssymb}
\usepackage[utf8]{inputenc}

\usepackage{microtype}
\usepackage{xstring}
\usepackage{booktabs}
\usepackage{color}
\usepackage{subfiles}
\usepackage{setspace}
\usepackage{caption}
\usepackage[textsize=tiny]{todonotes} 
\usepackage{subcaption}
\usepackage{enumitem}
\usepackage{amssymb}
\usepackage{pifont}

\usepackage{inconsolata}
\usepackage{amsmath}

\title{Incorporating External Knowledge and Goal Guidance for LLM-based Conversational Recommender Systems}
\author{Chuang Li$^{12}$, Yang Deng$^{1}$, Hengchang Hu$^{1}$, Min-Yen Kan$^{1}$, Haizhou Li$^{13}$ \\
        $^{1}$National University of Singapore\\
        $^{2}$NUS Graduate School for Integrative Sciences and Engineering \\
        $^{3}$Chinese University of Hong Kong, Shenzhen\\
        \texttt{\{lichuang, hengchanghu\}@u.nus.edu}\\
        \texttt{\{ydeng, kanmy, haizhou.li\}@nus.edu.sg}}

\begin{document}

\maketitle
\begin{abstract}


\newcommand{\model}[0]{ChatCRS }
This paper aims to efficiently enable large language models (LLMs) to use \textit{external knowledge} and \textit{goal guidance} in conversational recommender system (CRS) tasks. 
Advanced LLMs (\textit{e.g.}, ChatGPT) are limited in domain-specific CRS tasks for 1) generating grounded responses with recommendation-oriented knowledge,
or 2) proactively leading the conversations through different dialogue goals. 
In this work, we first analyze those limitations through a comprehensive evaluation, showing the necessity of external knowledge and goal guidance which contribute significantly to the recommendation accuracy and language quality. 
In light of this finding, we propose a novel \model framework to decompose the complex CRS task into several sub-tasks through the implementation of 1) a knowledge retrieval agent using a tool-augmented approach to reason over external Knowledge Bases and 2) a goal-planning agent for dialogue goal prediction. 
Experimental results on two multi-goal CRS datasets reveal that \model\ sets new state-of-the-art benchmarks, improving language quality of informativeness by 17\% and proactivity by 27\%, and achieving a tenfold enhancement in recommendation accuracy\footnote{Our code is publicly available at \href{https://github.com/lichuangnus/ChatCRS}{Git4ChatCRS}}.


\end{abstract}

\section{Introduction}






Conversational recommender system (CRS) integrates conversational and recommendation system (RS) technologies, naturally planning and proactively leading the conversations from non-recommendation goals (e.g., \textit{``chitchat''} or \textit{``question answering''}) to recommendation-related goals (e.g., \textit{``movie recommendation};  \citealp{A-survey-jannach, baseline_MGCG}). Compared with traditional RS, CRS highlights the multi-round interactions between users and systems using natural language. 
Besides the \textbf{recommendation task} evaluated by the recommendation accuracy as in RS, CRS also focuses on multi-round interactions in \textbf{response generation tasks} including asking questions, responding to user utterances or balancing recommendation versus conversation \cite{li2023conversation}.

\begin{figure}[!t]
\begin{center}
\includegraphics[width=8cm]{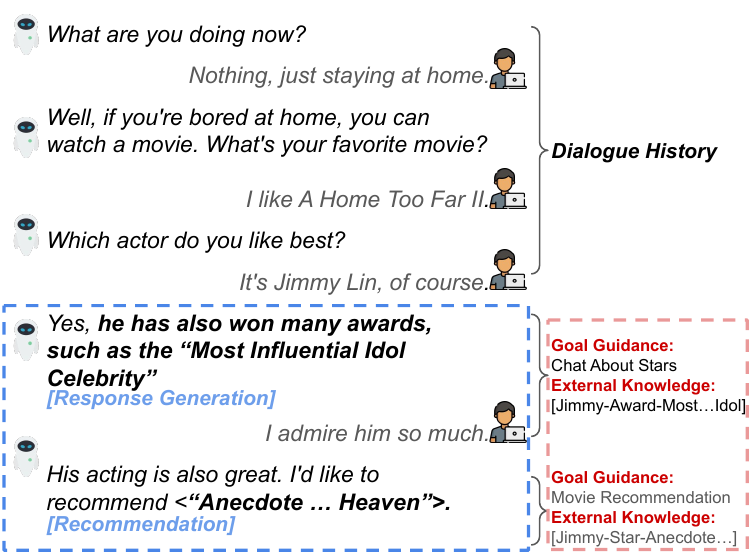}
\caption{An example of CRS tasks with external knowledge and goal guidance. ({\color{blue}Blue}: CRS tasks; {\color{red}Red}: External Knowledge and Goal Guidance)}

\label{example}
\end{center}
\vspace{-0.3cm}
\end{figure}
Large language models (LLMs; e.g., ChatGPT) that are significantly more proficient in response generation show great potential in CRS applications. However, current research concentrates on evaluating only their recommendation capability \cite{LLM_competitive_zero-shot, LLM_rec}. Even though LLMs demonstrate a competitive zero-shot recommendation proficiency, their recommendation performance primarily depends on content-based information (internal knowledge) and exhibits sensitivity towards demographic data \cite{he2023large, LLM_competitive_zero-shot}. Specifically, LLMs excel in domains with ample internal knowledge (e.g., English movies). 
However, in domains with scarce internal knowledge (e.g., Chinese movies\footnote{The Chinese movie domain encompasses CRS datasets originally sourced from Chinese movie websites, featuring both Chinese and international films.}), we found through our empirical analysis (\S~\ref{EA}) that their recommendation performance notably diminishes. Such limitation of LLM-based CRS motivates exploring solutions from prior CRS research to enhance domain coverage and task performance.

Prior work on CRS has employed general language models (LMs; e.g., {DialoGPT}) as the base architecture, but bridged the gap to domain-specific CRS tasks by incorporating external knowledge and goal guidance \cite{RecInDial, baseline_MGCG}. Inspired by this approach, we conduct an empirical analysis on the DuRecDial dataset \cite{liu-etal-2021-durecdial2} to understand how external inputs\footnote{In this paper, we limit the scope of external inputs to external knowledge and goal guidance.} can efficiently adapt LLMs in the experimented domain and enhance their performance on both recommendation and response generation tasks. 

Our analysis results (\S~\ref{EA}) reveal that despite their strong language abilities, LLMs exhibit notable limitations when directly applied to CRS tasks without external inputs in the Chinese movie domain. For example, lacking domain-specific knowledge (\textit{``Jimmy's Award''}) hinders the generation of pertinent responses, while the absence of explicit goals (\textit{``recommendation''}) leads to unproductive conversational turns (Figure~\ref{example}). Identifying and mitigating such constraints is crucial for developing effective LLM-based CRS \cite{li2023conversation}.

\begin{figure*}[!ht]

\includegraphics[width= \textwidth]{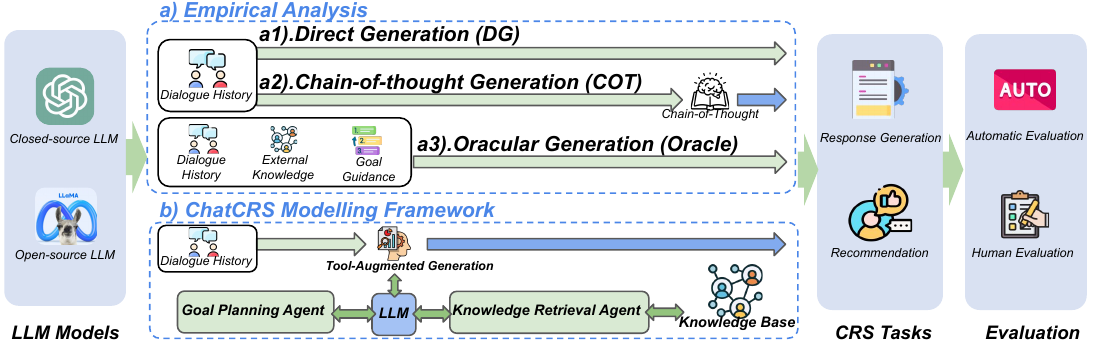} 
\caption{a) Empirical analysis of LLMs in CRS tasks with DG, COT\& Oracle; b) System design of ChatCRS framework using LLMs as a conversational agent to control the goal planning and knowledge retrieval agents.}
\label{system}
\end{figure*}

Motivated by the empirical evidence that external inputs can significantly boost LLM performance on both CRS tasks, we propose a novel \textbf{ChatCRS} framework. It decomposes the overall CRS problem into sub-components handled by specialized agents for knowledge retrieval and goal planning, all managed by a core LLM-based conversational agent. This design enhances the framework's flexibility, allowing it to work with different LLM models without additional fine-tuning while capturing the benefits of external inputs (Figure~\ref{system}b). Our contributions can be summarised as:

\begin{itemize}[leftmargin=*]
\setlength\itemsep{0.01cm}
    \item We present the first comprehensive evaluation of LLMs on both CRS tasks, including response generation and recommendation, and underscore the challenges in LLM-based CRS. 
     \item {We propose the ChatCRS framework as the first knowledge-grounded and goal-directed LLM-based CRS using LLMs as conversational agents.}
     \item {Experimental findings validate the efficacy and efficiency of ChatCRS in both CRS tasks. Furthermore, our analysis elucidates how external inputs contribute to LLM-based CRS.}
\end{itemize}
\vspace{-0.3cm}



\section{Related Work} \label{related}

\textbf{Attribute-based/Conversational approaches in CRS.} Existing research in CRS has been categorized into two approaches \cite{gao_advances_2021, li2023conversation}: 1) \textit{attribute-based approaches}, where the system and users exchange item attributes without conversation \cite{SAUR, EAR}, and 2) \textit{conversational approaches}, where the system interacts users through natural language \cite{li2018conversational,  uniCRS, baseline_TPNet}.

\vspace{0.8mm}

\noindent \textbf{LLM-based CRS.} LLMs have shown promise in CRS applications as 1) zero-shot conversational recommenders with item-based \cite{LLM_evaluate, LLM_rec} or conversational inputs \cite{he2023large, LLM_competitive_zero-shot, LLM-evaluation}; 2) AI agents controlling pre-trained CRS or LMs for CRS tasks \cite{LLM-expert, LLM_e-commerce, CRS-agent:2023recommender}; and 3) user simulators evaluating interactive CRS systems \cite{evaluation-rethinking-evaluation, evaluation_user_simulator, evaluation-huang2024concept}. However, there is a lack of prior work integrating external inputs to improve LLM-based CRS models.

\vspace{0.8mm}
\noindent \textbf{Multi-agent and tool-augmented LLMs.} LLMs, as conversational agents, can actively pursue specific goals through multi-agent task decomposition and tool augmentation \cite{RecAgent}. This involves delegating subtasks to specialized agents and invoking external tools like knowledge retrieval, enhancing LLMs' reasoning abilities and knowledge coverage \cite{agent1, agent2, gpt4tools, structgpt}.

In our work, we focus on the conversational approach, jointly evaluating CRS on both recommendation and response generation tasks \cite{ baseline_TPNet, li2023conversation, uniCRS}.  Unlike existing methods, ChatCRS uniquely combines goal planning and tool-augmented knowledge retrieval agents within a unified framework. This leverages LLMs' innate language and reasoning capabilities without requiring extensive fine-tuning.

\section{Preliminary: Empirical Analysis} \label{EA}
\label{s:pea}
We consider the CRS scenario where a system $system$ interacts with a user $u$. Each dialogue contains $T$ conversation turns with user and system utterances, denoted as $C = $\{$s_j^{system}, s_j^{u} $\}$^T_{j=1}$. The target function for CRS is expressed in two parts: given the dialogue history $C_j$ of the past $j^{th}$ turns, it generates 1) the recommendation of item $i$ and 2) the next system response $s_{j+1}^{system}$. 
In some methods, knowledge $K$ is given as an external input to facilitate both the recommendation and response generation tasks while dialogue goals $G$ only facilitate the response generation task due to the fixed ``recommendation'' goals in the recommendation task.
Given the user's contextual history $C_j$,  $system$ generates recommendation results $i$ and system response $s_{j+1}^{system}$ in Eq.~\ref{function}.
\begin{equation}
\label{function}
y^* = \prod\nolimits_{j=1}^T P_\theta~(i, s_{j+1}^{system}|~C_j, K, G ) 
\end{equation}


\subsection{Empirical Analysis Approaches} 


  

Building on the advancements of LLMs over general LMs in language generation and reasoning, we explore their inherent response generation and recommendation capabilities, with and without external knowledge or goal guidance. Our analysis comprises three settings, as shown in Figure~\ref{system}a: 
\begin{itemize}[leftmargin=*]
\setlength\itemsep{0.01cm}
    \item \textbf{\textit{Direct Generation (DG).}} LLMs directly generate system responses and recommendations without any external inputs (Figure~\ref{ICL}a). 
    \item \textbf{\textit{Chain-of-thought Generation (COT).}} LLMs internally reason their built-in knowledge and goal-planning scheme for both CRS tasks (Figure~\ref{ICL}b).
    \item \textbf{\textit{Oracular Generation (Oracle).}} LLMs leverage gold-standard external knowledge and dialogue goals to enhance performance in both CRS tasks, providing an upper bound (Figure~\ref{ICL}c).
\end{itemize}

 Additionally, we conduct an ablation study of different knowledge types on both CRS tasks by analyzing 1) factual knowledge, referring to general facts about entities and expressed as single triple (e.g., \textit{[Jiong--Star sign--Taurus]}), and 2) item-based knowledge, related to recommended items and expressed as multiple triples (e.g., \textit{[Cecilia--Star in--<movie 1, movie 2, ..., movie n>]}).
Our primary experimental approach utilizes in-context learning (ICL) on the \textit{DuRecDial} dataset \cite{liu-etal-2021-durecdial2}. Figure~\ref{ICL} provides an overview of the ICL prompts, with examples detailed in Appendix~\ref{Prompt} and experiments detailed in \S~\ref{exp}. For response generation, we evaluate content preservation ($bleu$-$n$, $F1$) and diversity ($dist$-$n$) with knowledge and goal prediction accuracy. For recommendation, we evaluate top-K ranking accuracy ($NDCG@k, MRR@k$).



\begingroup
\setlength{\tabcolsep}{4.6pt} 
\renewcommand{\arraystretch}{0.9} 

\begin{table}[t!]
\small
\centering
\begin{tabular}{cccc}
\toprule
$LLM$& $Task$  & $NDCG@10/50$ & $MRR@10/50$ \\
\midrule
\multirow{3}{*}{\rotatebox[origin=c]{0}{\textbf{\tiny ChatGPT}}}& DG &  0.024/0.035 & 0.018/0.020 \\
{}&COT-K  & 0.046/0.063 & 0.040/0.043  \\
{}&Oracle-K & \color{red}\textbf{0.617/0.624} & \color{red}\textbf{0.613/0.614} \\ \midrule
 \multirow{3}{*}{\rotatebox[origin=c]{0}{\textbf{\tiny LLaMA-7b}}}&DG & 0.013/0.020 & 0.010/0.010  \\
{}&COT-K  & 0.021/0.029 & 0.018/0.020  \\
{}&Oracle-K  & \color{red}\textbf{0.386/0.422} & \color{red}\textbf{0.366/0.370}  \\\midrule
\multirow{3}{*}{\rotatebox[origin=c]{0}{\textbf{\tiny LLaMA-13b}}}&DG & 0.027/0.031 & 0.024/0.024  \\ 
{}&COT-K  & 0.037/0.040 & 0.035/0.036  \\
{}&Oracle-K  & \color{red}{\textbf{0.724/0.734}}& \color{red}{\textbf{0.698/0.699}}  \\
\bottomrule
\end{tabular}
\caption{Empirical analysis for recommendation task in DuRecDial dataset ($K$: Knowledge; {\color{red}\textbf{Red}}: Best result).}
\label{table:EA_REC}
\end{table}
\endgroup

\begingroup
\setlength{\tabcolsep}{10pt} 
\renewcommand{\arraystretch}{0.9} 

\begingroup
\setlength{\tabcolsep}{12pt} 
\begin{table*}[ht]
\small
\tabcolsep=3mm
\centering
\begin{tabular}{cccccccccc}
\toprule
 $LLM$ & $Approach$ & $K/G$  & $bleu1$ & $bleu2$ & $bleu$ & $dist1$ & $dist2$ & $F1$ & $Acc_{G/K}$\\
\midrule
\multirow{7}{*}{\rotatebox[origin=c]{90}{\textbf{\large ChatGPT}}} & \textbf{DG}&   & 0.448 & 0.322& 0.161 & 0.330 & 0.814 & 0.522 & - \\\cmidrule(lr){2-9} 
 {}& \multirow{2}{*}{\textbf{COT}}&G  & 0.397 & 0.294 & 0.155& 0.294 & 0.779 & 0.499 & \underline{\textbf{0.587}} \\
{}& {}&K  & 0.467 & 0.323 & 0.156& 0.396 & 0.836 &0.474
& \underline{\textbf{0.095}} \\\cmidrule(lr){2-9} 
{}& \multirow{3}{*}{\textbf{Oracle}}&G   & 0.429 & 0.319& 0.172 & 0.315 & 0.796 & 0.519 & - \\
 {}& {}&K   & \color{red}\underline{\textbf{0.497}} & \color{red}\underline{\textbf{0.389}}& \color{red}\underline{\textbf{0.258}} & \color{red}\textbf{0.411}& \color{red}\underline{\textbf{0.843}}& 0.488 & - \\
 {}& {}&BOTH   & 0.428 & 0.341& 0.226 & 0.307 & 0.784 & \color{red}\textbf{0.525} & - \\
 \midrule
 \multirow{7}{*}{\rotatebox[origin=c]{90}{\textbf{\large  LLaMA-7b}}} & \textbf{DG}&  & 0.417 & 0.296 & 0.145 & 0.389 & 0.813 & 0.495 & - \\\cmidrule(lr){2-9} 
 {} & \multirow{2}{*}{\textbf{COT}}&G   & 0.418 & 0.293& 0.142 & 0.417 & 0.827 & 0.484 & 0.215 \\
 {} & {}&K   & 0.333 & 0.238 & 0.112& 0.320 & 0.762 & 0.455 & 0.026 \\
 \cmidrule(lr){2-9} 
 {} & \multirow{3}{*}{\textbf{Oracle}}&G  &  \color{red}\textbf{0.450} & \color{red}\textbf{0.322} &0.164& \color{red}\underline{\textbf{0.431}} & \color{red}{\textbf{0.834}} & \color{red}\textbf{0.504} & - \\
 {} & {}&K   & 0.359 & 0.270 & 0.154& 0.328 & 0.762 & 0.473 & - \\
 {}& {}&BOTH   & 0.425 & 0.320& \color{red}\textbf{0.187}& 0.412 & 0.807 & 0.492 & - \\\midrule
\multirow{7}{*}{\rotatebox[origin=c]{90}{\textbf{\large  LLaMA-13b}}} & \textbf{DG}&   & 0.418 & 0.303& 0.153 & 0.312 & 0.786 & 0.507 & - \\\cmidrule(lr){2-9} 
 {} &  \multirow{2}{*}{\textbf{COT}}&G   & 0.463 & 0.332 & 0.172& 0.348 & 0.816 & 0.528& 0.402 \\
 {} & {}&K   & 0.358 & 0.260 & 0.129& 0.276 & 0.755 & 0.473 & 0.023 \\\cmidrule(lr){2-9} 
{} & \multirow{3}{*}{\textbf{Oracle}}&G   & \color{red}{\textbf{0.494}} & \color{red}\textbf{0.361} & 0.197& \color{red}\textbf{0.373} & \color{red}\textbf{0.825}& \color{red}\underline{\textbf{0.543}} & - \\
 {} & {}&K   & 0.379 & 0.296 & 0.188& 0.278 & 0.754 & 0.495 & - \\
{}& {}&BOTH &  {0.460} & {0.357} &\color{red}{\textbf{0.229}}& {0.350} & {0.803}  & {0.539} & {-} \\
\bottomrule

\end{tabular}
\caption{Empirical analysis for response generation task in DuRecDial dataset ($K/G$: Knowledge or goal; $Acc_{G/K}$: Accuracy of knowledge or goal predictions; {\color{red}\textbf{Red}}: Best result for each model; \underline{Underline}: Best results for all).}
\label{table:EA_CRS}
\end{table*}
\endgroup

\subsection{Empirical Analysis Findings}
We summarize our three main findings given the results of the response generation and recommendation tasks shown in Tables~\ref{table:EA_REC} and~\ref{table:EA_CRS}.

\vspace{0.8mm}
\noindent \textit{\textbf{Finding 1}: The Necessity of External Inputs in LLM-based CRS.} 
Integrating external inputs significantly enhances performance across all LLM-based CRS tasks (Oracle), underscoring the insufficiency of LLMs alone as effective CRS tools and highlighting the indispensable role of external inputs. Remarkably, the Oracle approach yields over a tenfold improvement in recommendation tasks with only external knowledge compared to DG and COT methods, as the dialogue goal is fixed as ``recommendation'' (Table~\ref{table:EA_REC}). Although utilizing internal knowledge and goal guidance (COT) marginally benefits both tasks, we see in Table~\ref{table:EA_CRS} for the response generation task that the low accuracy of internal predictions adversely affects performance.

\begingroup
\setlength{\tabcolsep}{5pt} 
\renewcommand{\arraystretch}{1} 

\vspace{0.8mm}
\noindent \textit{\textbf{Finding 2}: Improved Internal Knowledge or Goal Planning Capability in Advanced LLMs.}
Table~\ref{table:EA_CRS} reveals that the performance of Chain-of-Thought (COT) by a larger LLM (LLaMA-13b) is comparable to oracular performance of a smaller LLM (LLaMA-7b). This suggests that the intrinsic knowledge and goal-setting capabilities of more sophisticated LLMs can match or exceed the benefits derived from external inputs used by their less advanced counterparts. Nonetheless, such internal knowledge or goal planning schemes are still insufficient for CRS in domain-specific tasks while the integration of more accurate knowledge and goal guidance (Oracle) continues to enhance performance to state-of-the-art (SOTA) outcomes.

\vspace{0.8mm}
\noindent \textit{\textbf{Finding 3}: Both factual and item-based knowledge jointly improve LLM performance on domain-specific CRS tasks.}
As shown in Table~\ref{table:EA_KKK}, integrating both factual and item-based knowledge yields performance gains for LLMs on both response generation and recommendation tasks. Our analysis suggests that even though a certain type of knowledge may not directly benefit a CRS task (e.g., factual knowledge may not contain the target items for the recommendation task), it can still benefit LLMs by associating unknown entities with their internal knowledge, thereby adapting the universally pre-trained LLMs to task-specific domains more effectively. Consequently, we leverage both types of knowledge jointly in our ChatCRS framework.

\begingroup
\setlength{\tabcolsep}{6pt} 
\renewcommand{\arraystretch}{1} 
\begin{table}[t!]
\small
    \centering
    \begin{tabular}{lcc}
    \toprule
    \multicolumn{3}{c}{Response Generation Task}\\ \midrule
        $Knowledge$ & $bleu1/2/F1$ & $dist1/2$ \\
        \midrule
      
        \textbf{\textit{Both Types}} &\textbf{0.497/0.389/0.488} & \textbf{0.411/0.843} \\
        \textbf{\textit{ -w/o Factual*}} &{0.407/0.296/0.456}& {0.273/0.719} \\
        \textbf{\textit{ -w/o Item-based*}} &{0.427/0.331/0.487}& {0.277/0.733}  \\
        \toprule
        \multicolumn{3}{c}{Recommendation Task}\\ \midrule
        $Knowledge$ & $NDCG@10/50$ & $MRR@10/50$ \\
        \midrule
      
        \textbf{\textit{Both Types}} &\textbf{0.617/0.624} & \textbf{0.613/0.614} \\
        \textbf{\textit{ -w/o Factual*}} &{0.272/0.290}& {0.264/0.267} \\
        \textbf{\textit{ -w/o Item-based*}} &{0.376/0.389}& {0.371/0.373}  \\
        
        \bottomrule
    \end{tabular}
    \caption{Ablation study for ChatGPT with different knowledge types in DuRecDial dataset. }
    
\label{table:EA_KKK}
\end{table}
\endgroup



\section{ChatCRS} \label{MF}

Our ChatCRS modelling framework has three components: 1) a knowledge retrieval agent, 2) a goal planning agent and 3) an LLM-based conversational agent (Figure~\ref{system}b). Given a complex CRS task, an LLM-based conversational agent first decomposes it into subtasks managed by knowledge retrieval or goal-planning agents. The retrieved knowledge or predicted goal from each agent is incorporated into the ICL prompt to instruct LLMs to generate CRS responses or recommendations.

\subsection{Knowledge Retrieval agent} \label{KR}

Our analysis reveals that integrating both factual and item-based knowledge can significantly boost the performance of LLM-based CRS. However, knowledge-enhanced approaches for LLM-based CRS present unique challenges that have been relatively unexplored compared to prior \textit{training-based methods} in CRS or \textit{retrieval-augmented (RA) methods} in NLP  \cite{KG-LLM-1, KG-LLM-Retrieval}. 

Training-based methods, which train LMs to memorize or interpret knowledge representations through techniques like graph propagation, have been widely adopted in prior CRS research \cite{KG-enhanced, zhang2023variational}. However, such approaches are computationally infeasible for LLMs due to their input length constraints and training costs.
RA methods, which first collect evidence and then generate responses, face two key limitations in CRS \cite{manzoor_retrieval-conf, RAG-LLM}. First, without a clear query formulation in CRS, RA methods can only approximate results rather than retrieve the exact relevant knowledge \cite{RAG, RAG-fail}. Especially when multiple similar entries exist in the knowledge base (KB), precisely locating the accurate knowledge for CRS becomes challenging.
Second, RA methods retrieve knowledge relevant only to the current dialogue turn, whereas CRS requires planning for potential knowledge needs in future turns, differing from knowledge-based QA systems \cite{RAG-QA, structgpt}. For instance, when discussing a celebrity without a clear query (e.g.,  \textit{``I love Cecilia...''}), the system should anticipate retrieving relevant factual knowledge (e.g., \textit{``birth date''} or \textit{``star sign''}) or item-based knowledge (e.g., \textit{``acting movies''}) for subsequent response generation or recommendations, based on the user's likely interests.

To address this challenge, we employ a relation-based method which allows LLMs to flexibly plan and quickly retrieve relevant ``entity--relation--entity'' knowledge triples $K$ by traversing along the relations $R$ of mentioned entities $E$ \cite{open-dial-KG, structgpt}. Firstly, 
\noindent \textbf{entities} for each utterance is directly provided by extracting entities in the knowledge bases from the dialogue utterance \cite{zou_improving_2022}. 
\noindent \textbf{Relations} that are adjacent to entity $E$ from the KB are then extracted as candidate relations (denoted as $F1$) and LLMs are instructed to plan the knowledge retrieval by selecting the most pertinent relation $R^*$ given the dialogue history $C_j$. 
\noindent \textbf{Knowledge triples $K^*$} can finally be acquired using entity $E$ and predicted relation $R^*$ (denoted as $F2$). The process is formulated in  Figure~\ref{KG-a} and demonstrated with an example in Figure~\ref{KG-b}.  Given the dialogue utterance \textit{``I love Cecilia...''} and the extracted entity \textit{[Cecilia]}, the system first extracts all potential relations for \textit{[Cecilia]}, from which the LLM selects the most relevant relation, \textit{[Star in]}. The knowledge retrieval agent then fetches the complete knowledge triple \textit{[Cecilia--Star in--<movie 1, movie 2, ..., movie n>]}.


\begin{figure}[!t]
\begin{center}
\includegraphics[width=0.48\textwidth]{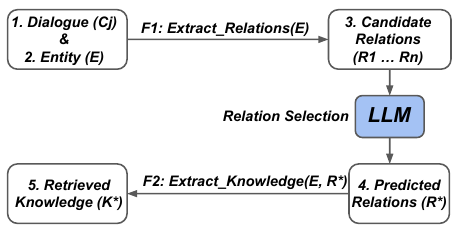}
\caption{Knowledge retrieval agent in ChatCRS.}
\label{KG-a}
\end{center}
\vspace{-0.3cm}
\end{figure}

\begingroup
\setlength{\tabcolsep}{10.5pt} 
\renewcommand{\arraystretch}{1} 
\begin{table*}[t!]
\small
\tabcolsep=3mm
\centering
\begin{tabular}{lcccccccccc}
\toprule
       \multirow{2}{*}{\textbf{Model}}& \multirow{2}{*}{\textbf{N-shot}}&\multicolumn{4}{c}{\textbf{DuRecDial}}& \multicolumn{4}{c}{\textbf{TG-Redial}}\\\cmidrule(lr){3-6} \cmidrule(lr){7-10} 
&&  $bleu1$ &  $bleu2$& $dist2$ & $F1$&  $bleu1$ &  $bleu2$& $dist2$ & $F1$ \\
\midrule
{MGCG}& $Full$& 0.362&0.252 & 0.081&0.420& NA & NA & NA& NA\\
{MGCG-G}&$ Full$& 0.382&0.274 & 0.214&0.435& NA & NA & NA& NA\\
{TPNet}& $Full$ &0.308&0.217 & 0.093&0.363& NA & NA & NA& NA\\
{UniMIND*} & $Full$&0.418&0.328 & 0.086 &0.484&{0.291}&0.070 & 0.200 & \textbf{0.328}\\
{ChatGPT}  & $3$ &0.448&0.322 & \textbf{0.814}  & 0.522&0.262&0.126 & \textbf{0.987}  & 0.266  \\
{{LLaMA}}  & $3$ &0.418&0.303 & 0.786 &  0.507& 0.205&0.096 & 0.970 & 0.247   \\
\textbf{ChatCRS}  & $3$ & \textbf{0.460}&\textbf{0.358}&0.803&\textbf{0.540}&\textbf{0.300}&\textbf{0.180} &\textbf{0.987} &0.317\\
\bottomrule

\end{tabular}
\caption{Results of response generation task on DuRecDial and TG-Redial datasets. (UniMIND*: Results from the ablation study in the original UniMIND paper.)}
\label{table:MF_CRS}
\end{table*}
\endgroup
\begingroup
\setlength{\tabcolsep}{10pt} 
\renewcommand{\arraystretch}{1} 
\begin{table*}[t!]
\small
\tabcolsep=3mm
\centering
\begin{tabular}{lccccc}
\toprule
 \multirow{2}{*}{\textbf{Model}}& \multirow{2}{*}{\textbf{N-shot}}&\multicolumn{2}{c}{\textbf{DuRecDial}}& \multicolumn{2}{c}{\textbf{TG-Redial}}\\\cmidrule(lr){3-4} \cmidrule(lr){5-6} 
 && $NDCG@10 / 50$ & $MRR@10 / 50$ & $NDCG@10/50$ & $MRR@10/50$ \\
\midrule
{SASRec} & $Full$& {0.369} / {0.413} & {0.307} / {0.317}& 0.009 / 0.018 & 0.005 / 0.007\\
{UniMIND} & $Full$& \textbf{0.599} / \textbf{0.610} & \textbf{0.592} / \textbf{0.594}& \textbf{0.031} / \textbf{0.050} & 0.024 / 0.028\\
{{ChatGPT}} &3& 0.024 / 0.035 & 0.018 / 0.020& 0.001 / 0.003 & 0.005 / 0.005 \\
{{LLaMA}} &3& 0.027 / 0.031 & 0.024 / 0.024 &  0.001 / 0.006 & 0.003 / 0.005\\ 
{\textbf{ChatCRS}}&3&  \textbf{0.549}  /  \textbf{0.553} & \textbf{0.543}  /  \textbf{0.543}&\textbf{0.031}  /  \textbf{0.033} & \textbf{0.082}  /  \textbf{0.083} \\ 
\bottomrule
\end{tabular}
\caption{Results of recommendation task on DuRecDial and TG-Redial datasets.}
\label{table:MF_REC}
\end{table*}
\endgroup

When there are multiple entities in one utterance, we perform the knowledge retrieval one by one and in the scenario where there are multiple item-based knowledge triples, we randomly selected a maximum of 50 item-based knowledge due to the limitations of input token length. 
We implement N-shot ICL to guide LLMs in choosing knowledge relations and we show the detailed ICL prompt and instruction with examples in Table~\ref{table: KR} (\S~\ref{AK}). 

\subsection{Goal Planning agent} 

Accurately predicting the dialogue goals is crucial for 1) proactive response generation and 2) balancing recommendations versus conversations in CRS. Utilizing goal annotations for each dialogue utterance from CRS datasets, we leverage an existing language model, adjusting it for goal generation by incorporating a Low-Rank Adapter (LoRA) approach \cite{LORA, qlora}. This method enables parameter-efficient fine-tuning by adjusting only the rank-decomposition matrices. For each dialogue history $C^k_j$ ($j$-$th$ turn in dialogue $k$; $j \in T$, $k\in N$), the LoRA model is trained to generate the dialogue goal $G^*$ for the next utterance using the prompt of dialogue history, optimizing the loss function in Eq~\ref{Goal-func} with $\theta$ representing the trainable parameters of LoRA. The detailed prompt and instructions are shown in Table~\ref{table: GP} (\S~\ref{AG}).
\begin{equation}
\label{Goal-func}
L_g = -\sum\nolimits^N_k~\sum\nolimits^T_j{\log P_{\theta}~(G^*|~C^k_j)} 
\end{equation}

\subsection{LLM-based Conversational Agent} 

In ChatCRS, the knowledge retrieval and goal-planning agents serve as essential tools for CRS tasks, while LLMs function as tool-augmented conversational agents that utilize these tools to accomplish primary CRS objectives. Upon receiving a new dialogue history $C_j$, the LLM-based conversational agent employs these tools to determine the dialogue goal $G^*$ and relevant knowledge $K^*$, which then instruct the generation of either a system response $s_{j+1}^{system}$ or an item recommendation $i$ through prompting scheme, as formulated in Eq~\ref{LLM-func}. The detailed ICL prompt can be found in \S~\ref{Prompt}.


\begin{equation}
\label{LLM-func}
i, ~s_{j+1}^{system} = LLM(~C_j, K^*, G^*)
\end{equation}

\section{Experiments} \label{exp}

\subsection{Experimental Setups}
\textbf{Datasets.} We conduct the experiments on two multi-goal Chinese CRS benchmark datasets a) DuRecDial \cite{liu-etal-2021-durecdial2} in English and Chinese, and b) TG-ReDial \cite{zhou_towards_2020_TGRedial} in Chinese (statistics in Table \ref{data}). Both datasets are annotated for goal guidance, while only DuRecDial contains knowledge annotation and an external KB--CNpedia \cite{zhou_c2-crs_2022} is used for TG-Redial.  
\newline\textbf{Baselines.} We compare our model with ChatGPT\footnote{OpenAI API: gpt-3.5-turbo-1106} and LLaMA-7b/13b \cite{llama2} in few-shot settings. We also compare fully-trained UniMIND \cite{uniCRS}, MGCG-G\cite{baseline_MGCG}, TPNet\cite{baseline_TPNet}, MGCG \cite{liu_towards_2020_DuRecDial} and SASRec \cite{kang_self-attentive_2018}, which are previous SOTA CRS and RS models and we summarise each baseline in \S~\ref{A1}.  
\newline\textbf{Automatic Evaluation.}\label{human} For response generation evaluation, we adopt $BLEU$, $F1$ for content preservation and $Dist$ for language diversity. For recommendation evaluation, we adopt $NDCG@k$ and $MRR@K$ to evaluate top K ranking accuracy. For the knowledge retrieval agent, we adopt Accuracy ($Acc$), Precision ($P$), Recall ($R$) and $F1$ to evaluate the accuracy of relation selection (\S~\ref{AK}).
\newline\textbf{Human Evaluation.} For human evaluation, we randomly sample 100 dialogues from DuRecDial, comparing the responses produced by UniMIND, ChatGPT, LLaMA-13b and ChatCRS. Three annotators are asked to score each generated response with \{0: poor, 1: ok, 2: good\} in terms of a) general language quality in (Flu)ency and (Coh)erence, and b) CRS-specific language qualities of (Info)rmativeness and (Pro)activity. Details of the process and criterion are discussed in \S~\ref{A2}.
\newline\textbf{Implementation Details.}
For both the CRS tasks in Empirical Analysis, we adopt N-shot ICL prompt settings on ChatGPT and LLaMA* \cite{ICL}, where $N$ examples from the training data are added to the ICL prompt. 
In modelling framework, 
for the goal planning agent, we adopt QLora as a parameter-efficient way to fine-tune LLaMA-7b \cite{qlora}. For the knowledge retrieval agent and LLM-based conversational agent, we adopt the same N-shot ICL approach on ChatGPT and LLaMA* \cite{structgpt}. Detailed experimental setups are discussed in \S~\ref{A1}.

\begingroup
\setlength{\tabcolsep}{6pt} 
\renewcommand{\arraystretch}{1} 
\begin{table}[t!]
\small
    \centering
    \begin{tabular}{llllll}
        \toprule
        \multirow{2}{*}{\textbf{Model}} & \multicolumn{2}{c}{\textbf{General}} & \multicolumn{2}{c}{\textbf{CRS-specific}}\\
        \cmidrule(lr){2-3} \cmidrule(lr){4-5} 
        {} & \textit{Flu} & \textit{Coh} & \textit{Info}  & \textit{Pro} & \textit{Avg.}\\
        \midrule
        UniMIND & 1.87& 1.69& 1.49& 1.32& 1.60 \\
        ChatGPT &  \textbf{1.98}& 1.80& 1.50& 1.30& 1.65\\
        LLaMA-13b &1.94&1.68& 1.21& 1.33& 1.49\\\midrule
        \textbf{\textit{ChatCRS}} &\textbf{1.99}& \textbf{1.85}&  \textbf{1.76}&  \textbf{1.69}&  \textbf{1.82}  \\
        \textbf{\textit{ -w/o K*}} &\textbf{2.00}& \textbf{1.87} & 1.49 $\downarrow$ & 1.62& 1.75\\
        \textbf{\textit{ -w/o G*}} &\textbf{1.99}& \textbf{1.85}& 1.72& 1.55 $\downarrow$& 1.78  \\
        
        \bottomrule
    \end{tabular}
    \caption{Human evaluation and ChatCRS ablations for language qualities of (Flu)ency, (Coh)erence, (Info)rmativeness and (Pro)activity on  DuRecDial ($K^*/G^*$: Knowledge retrieval or goal-planning agent).}
    \label{table: human}
\end{table}
\endgroup

\begingroup
\setlength{\tabcolsep}{5pt} 
\renewcommand{\arraystretch}{1} 
\begin{table}[t!]
\centering
\small
\begin{tabular}{lccccc}
\toprule
\multirow{2}{*}{\textbf{Model}}  & \multicolumn{5}{c}{\textbf{Knowledge}} \\
\cmidrule(lr){2-6} 
        {}&N-{shot} & Acc  & P  & R & F1  \\\midrule
TPNet & $Full$& NA & NA & NA & 0.402 \\
MGCG-G& $Full$& NA & 0.460 & 0.478 & 0.450 \\
ChatGPT & 3 &0.095&0.031&0.139&0.015 \\
LLaMA-13b& 3&0.023&0.001&0.001&0.001 \\
\textbf{ChatCRS} & 3&\textbf{0.560}& \textbf{0.583} & \textbf{0.594} & \textbf{0.553} \\

\bottomrule
\end{tabular}
\caption{Results for knowledge retrieval on DuRecDial.}
\label{table: know}
\end{table}
\endgroup


\subsection{Experimental Results} \label{Res}

\textbf{\textit{ChatCRS significantly improves LLM-based conversational systems for CRS tasks,}} outperforming SOTA baselines in response generation in both datasets, enhancing content preservation and language diversity (Table~\ref{table:MF_CRS}). ChatCRS sets new SOTA benchmarks on both datasets using 3-shot ICL prompts incorporating external inputs.
In recommendation tasks (Table~\ref{table:MF_REC}), LLM-based approaches lag behind full-data trained baselines due to insufficient in-domain knowledge. Remarkably, \textit{ChatCRS}, by harnessing external knowledge, achieves a tenfold increase in recommendation accuracy over existing LLM baselines on both datasets with ICL, without full-data fine-tuning.

\textbf{\textit{Human evaluation highlights ChatCRS's enhancement in CRS-specific language quality.}} 
Table~\ref{table: human} shows the human evaluation and ablation results. ChatCRS outperforms baseline models in both general and CRS-specific language qualities. While all LLM-based approaches uniformly exceed the general LM baseline (UniMIND) in general language quality, ChatCRS notably enhances coherence through its goal guidance feature, enabling response generation more aligned with the dialogue goal. Significant enhancements in CRS-specific language quality, particularly in informativeness and proactivity, underscore the value of integrating external knowledge and goals. Ablation studies, removing either knowledge retrieval or goal planning agent, demonstrate a decline in scores for informativeness and proactivity respectively, confirming the efficacy of both external inputs for CRS-specific language quality. 
\begin{figure}[!t]
\begin{center}
\includegraphics[width=7.5cm]{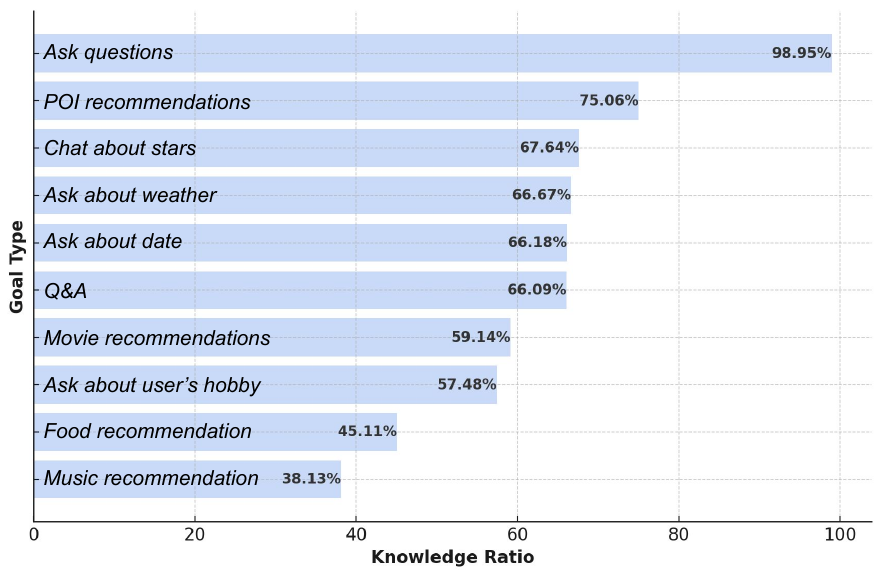}
\caption{Knowledge ratio for each goal type on DuRecDial. (X-axis: Knowledge Ratio ; Y-axis: Goal type)}
\label{ratio}
\end{center}
\end{figure}
\subsection{Detailed Discussion}

\textbf{CRS datasets typically contain a huge volume of knowledge.} By analyzing dialogues from the DuRecDial datasets, categorized by goal types, we calculated a ``Knowledge Ratio''  dividing the number of utterances with annotated knowledge $N_{K, G}$ by total number of utterances $N_{G}$ in each goal type (Eq~\ref{knowledge_ratio}) to measure the necessity of relevant knowledge in CRS task completion. Our analysis, depicted in Figure~\ref{ratio}, shows that recommendation tasks rank highly in terms of knowledge necessity, with ``POI recommendation'' dialogues requiring pertinent knowledge in 75\% of cases.
\begin{equation}
\label{knowledge_ratio}
    \text{Knowledge Ratio (KR)}_G = \frac{N_{K,G}}{N_G}
\end{equation}

\begin{table*}[t!]
\small
\begin{tabular}{ p{5cm}|p{9.5cm}} 
 \toprule{\textbf{Case Study 1: Factual Knowledge QA}}&{\textbf{Case Study 2: Proactive Response Generation}} \\\midrule
 \textbf{Dialogue history (Dial\_ID 39-1):} \newline {[user]: Do you know Jiong He's zodiac sign ?}
 &  \textbf{Dialogue history (Dial\_ID 89-14):}\newline {[user]: I like Cecilia Cheung very much. Her acting is very good.}\newline {\textbf{.  . .}}\newline {[system]: Yeah, have you seen Cecilia Cheung's One Night in Mongkok?}\newline {[user]: I've seen it. I don't want to see it again.}
 \\\midrule

  {\textbf{System Responses:} }
  \newline {\textbf{[\ding{51}]~UniMIND}: Of course, Taurus}
 \newline {\textbf{{[\color{red}\ding{55} }]~ChatGPT}: Jiong He's zodiac sign  is Cancer.}
 \newline {\textbf{{[\color{red}\ding{55} }]~LLaMA}: Jiong He's zodiac sign is Aries.}
  \newline \textbf{[\ding{51}]~ChatCRS}: {Yes, I do! Jiong He's zodiac sign is Taurus. \newline \textbf{\textit{[Retrieved Knowledge: \newline (`Jiong He', `zodiac sign ', `Taurus')]}}}
  & {\textbf{System Responses:} }
  \newline {\textbf{[\ding{51}]~UniMIND}: How about <Left Right...Destiny>, a very touching movie.}
 \newline {\textbf{[\ding{51}]~ChatGPT}: How about trying her other film, <Lost in Time>, a touching romance that showcases her acting skills?}
 \newline {\textbf{{[\color{red}\ding{55} }]~LLaMA}: Maybe you can try it again, you may find a different feeling.}
  \newline {\textbf{[\ding{51}]~ChatCRS}: If you haven't seen it, you should watch <Left Right...Destiny>, it's Cecilia Cheung's new movie this year. She played the role of a taxi driver, very cool! Have you seen it? \newline \textbf{\textit{[Predicted Goal: Movie Recommendation]}}}\\\bottomrule
\end{tabular}
\caption{Case study for ChatCRS with baselines.}
\label{case}
\end{table*}

\textbf{Internal knowledge of LLMs is insufficient for handling domain-specific CRS tasks}, which is reaffirmed by the analysis of knowledge retrieval accuracy in Table~\ref{table: know}.
Contrasting with traditional RS which relies on user data for collaborative recommendations, CRS mainly depends on context/content-based recommendation \cite{he2023large}. This shift highlights the limitations of LLMs in harnessing internal knowledge. ChatCRS overcomes these limitations by interfacing LLMs to plan and reason over external KBs through entities and relations. 
Therefore, it largely improves the recommendation accuracy, outperforming the training-based approach using full data. Given the limitations in LLM-based CRS tasks, \cite{KG-LLM-1, KG-LLM-Retrieval}, we anticipate future studies to further explore such approaches in CRS.

\textbf{Factual knowledge guides the response generation process, mitigating the risks of generating implausible or inconsistent responses}. The ``Asking questions'' goal type which has the highest knowledge ratio, demonstrates the advantage of leveraging external knowledge in answering factual questions like \textit{``the zodiac sign of an Asian celebrity''} (Table~\ref{case}). Standard LLMs produce responses with fabricated content, but ChatCRS accurately retrieves and integrates external knowledge, ensuring factual and informative responses. 


\textbf{{Goal guidance contributes more to the linguistic quality of CRS by managing the dialogue flow}.} We examine the goal planning proficiency of ChatCRS by showcasing the results of goal predictions of the top 5 goal types in each dataset (Figure~\ref{GOALS}). DuRecDial dataset shows better balances among recommendation and non-recommendation goals, which exactly aligns with the real-world scenarios \cite{INSPIRED-shirley}. However, the TG-Redial dataset contains more recommendation-related goals and multi-goal utterances, making the goal predictions more challenging. The detailed goal planning accuracy is discussed in \S~\ref{A-goal}. 

\textbf{Dialogue goals guide LLMs towards a proactive conversational recommender.} For a clearer understanding, we present a scenario in Table~\ref{case} where a CRS seamlessly transitions between ``asking questions'' and ``movie recommendation'', illustrating how accurate goal direction boosts interaction relevance and efficacy. Specifically, if a recommendation does not succeed, ChatCRS will adeptly pose further questions to refine subsequent recommendation responses while LLMs may keep outputting wrong recommendations, creating unproductive dialogue turns. This further emphasizes the challenges of conversational approaches in CRS, where the system needs to proactively lead the dialogue from non-recommendation goals to approach the users' interests for certain items or responses \cite{baseline_MGCG}, and underscores the goal guidance in fostering proactive engagement in CRS.



\section{Conclusion} \label{con}

This paper conducts an empirical investigation into the LLM-based CRS for domain-specific applications in the Chinese movie domain,  emphasizing the insufficiency of LLMs in domain-specific CRS tasks and the necessity of integrating external knowledge and goal guidance. We introduce ChatCRS, a novel framework that employs a unified agent-based approach to more effectively incorporate these external inputs. Our experimental findings highlight improvements over existing benchmarks, corroborated by both automatic and human evaluation. ChatCRS marks a pivotal advancement in CRS research, fostering a paradigm where complex problems are decomposed into subtasks managed by agents, which maximizes the inherent capabilities of LLMs and their domain-specific adaptability in CRS applications.



\section*{Limitations} \label{Limitations}
This research explores the application of few-shot learning and parameter-efficient techniques with large language models (LLMs) for generating responses and making recommendations, circumventing the need for the extensive fine-tuning these models usually require. Due to budget and computational constraints, our study is limited to in-context learning with economically viable, smaller-scale closed-source LLMs like ChatGPT, and open-source models such as LLaMA-7b and -13b.

A significant challenge encountered in this study is the scarcity of datasets with adequate annotations for knowledge and goal-oriented guidance for each dialogue turn. This limitation hampers the development of conversational models capable of effectively understanding and navigating dialogue. It is anticipated that future datasets will overcome this shortfall by providing detailed annotations, thereby greatly improving conversational models' ability to comprehend and steer conversations.

\section*{Ethic Concerns} \label{Ethic}
The ethical considerations for our study involving human evaluation (\S~\ref{human}) have been addressed through the attainment of an IRB Exemption for the evaluation components involving human subjects. The datasets utilized in our research are accessible to the public \cite{liu-etal-2021-durecdial2, zhou_towards_2020_TGRedial}, and the methodology employed for annotation adheres to a double-blind procedure (\S~\ref{human}). Additionally, annotators receive compensation at a rate of \$15 per hour, which is reflective of the actual hours worked.


\bibliography{ChatCRS}
\bibliographystyle{acl_natbib}

\appendix

\section{Appendix}
\begin{figure*}
\includegraphics[width= \textwidth]{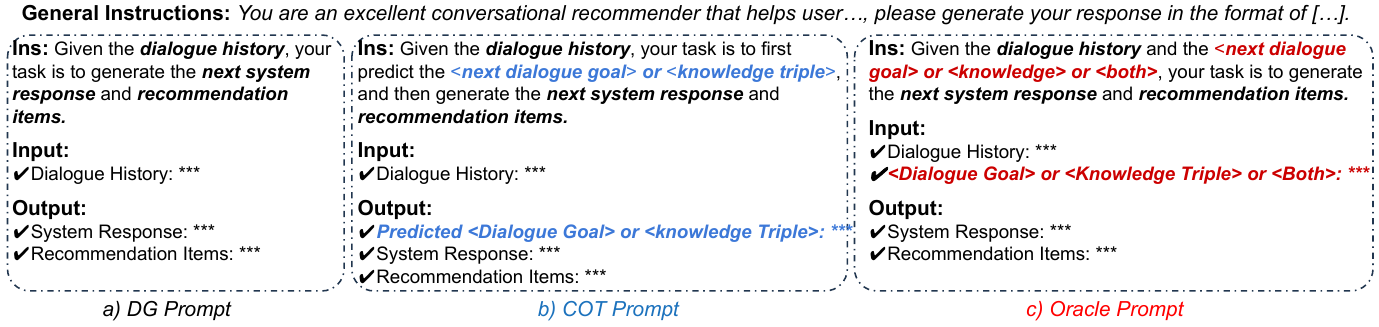} 
\caption{ICL prompt design for empirical analysis, detailed examples are shown in Appendix~\ref{Prompt}. 
}
\label{ICL}
\end{figure*}

\begin{figure*}[!t]
     \centering
     \begin{subfigure}{0.49\textwidth}
         \centering
         \includegraphics[width=\textwidth]{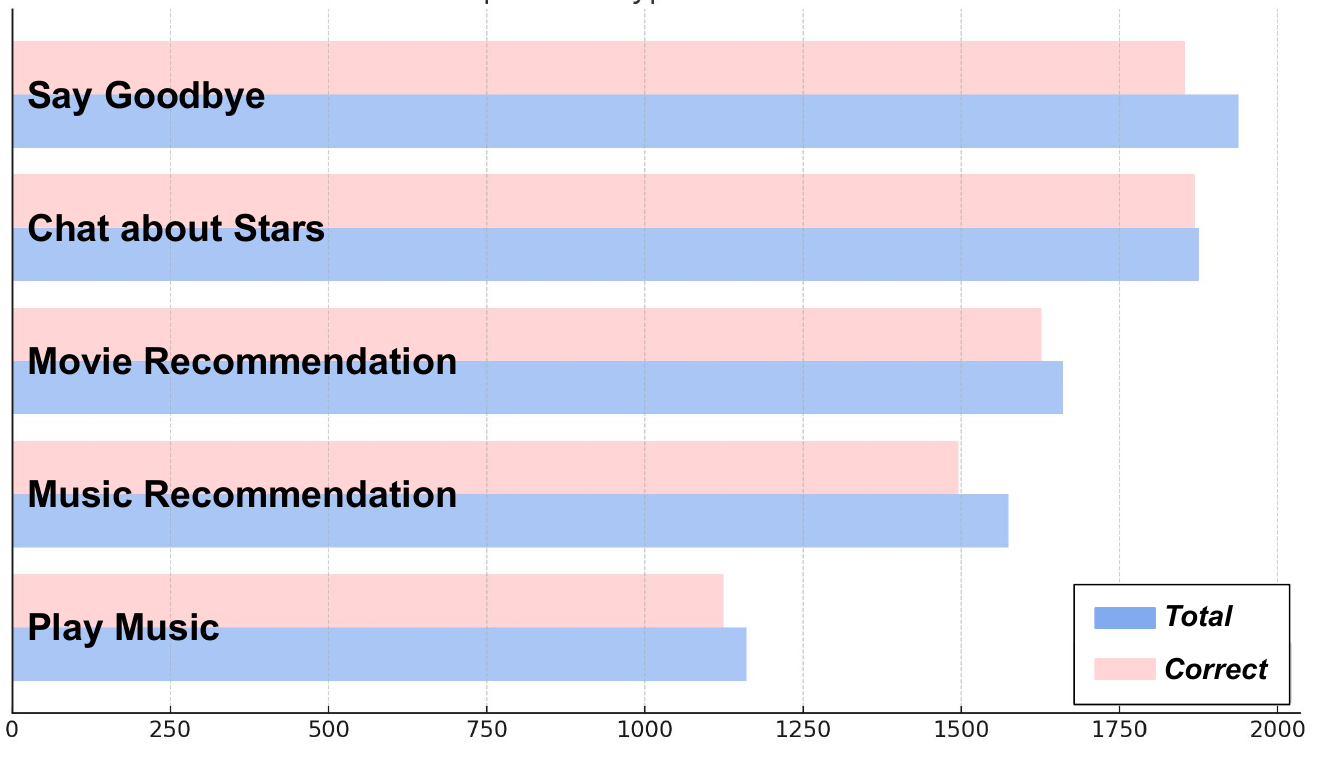}
         \caption{Results of goal predictions for DuRecDial dataset.}
         \label{KG-a1}
     \end{subfigure}%
     \begin{subfigure}{0.48\textwidth}
         \centering
         \includegraphics[width=\textwidth]{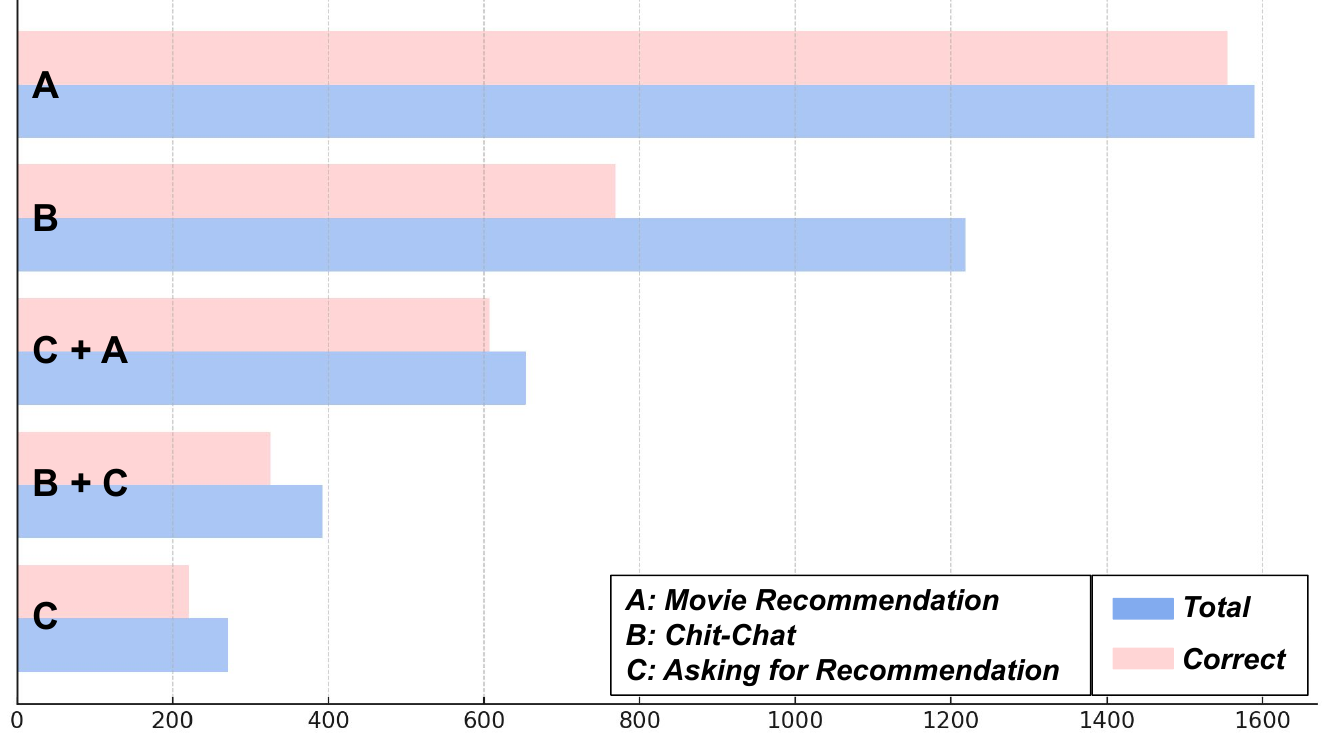}
         \caption{Results of goal predictions for TG-Redial datasets.}
         \label{KG-b1}
     \end{subfigure}
     \hfill
        \caption{Results of ChatCRS goal predictions with different goal types.}
        \label{GOALS}
\end{figure*}

\subsection{ICL Prompt for Empirical Analysis}\label{Prompt}
In Section \S~\ref{EA}, we examine the capabilities of Large Language Models (LLMs) through various empirical analysis methods: Direct Generation (DG), Chain-of-Thought Generation (COT), and Oracular Generation (Oracle). These approaches assess both the intrinsic abilities of LLMs and their performance when augmented with internal or external knowledge or goal directives. The description and testing objective of each empirical analysis methods is shown as follows:

\begin{itemize}[leftmargin=*]
\setlength\itemsep{0.01cm}
    \item \textbf{\textit{Direct Generation (DG).}} Utilizing dialogue history, DG produces system responses and recommendations to assess the model's inherent capabilities in two CRS tasks (Figure~\ref{ICL}a). 
    \item \textbf{\textit{Chain-of-thought Generation (COT).}} With dialogue history as input, COT generates knowledge or goal predictions before generating system responses and recommendations. We evaluate the model's efficacy using only its internal knowledge and goal-setting mechanisms (Figure~\ref{ICL}b).
    \item \textbf{\textit{Oracular Generation (Oracle).}} By incorporating dialogue history,  and ground truth external knowledge and goal guidance, Oracle generates system responses and recommendations. This yields an upper-bound, potential performance of LLMs in CRS tasks (Figure~\ref{ICL}c).
\end{itemize}

We provide the ICL prompt design in Table~\ref{ICL} and sample instructions within the prompts in Table~\ref{p:ins}. Furthermore, we detail the actual input-output examples presented in Table~\ref{p:out}.

\subsection{Detailed Knowledge Retrieval Agent} \label{AK}
For the knowledge retrieval agent in ChatCRS, we adopt a 3-shot ICL approach to guide LLMs in planning and selecting the best knowledge for the next utterance by traversing through the relations of the entity, as discussed in \S~\ref{KR}. For each dialogue history, we first extract the entity in the utterance from the knowledge base and then extract all the candidate relations of the entity from the knowledge base. Given the entity, candidate relations and dialogue history, we use instructions to prompt LLMs in planning and select the relations relevant to the knowledge or topics in the next utterance, as shown in Figure~\ref{KG-b}. We use 3-shot ICL for our experiment in knowledge retrieval with examples of 3 dialogue histories ($C_j$) randomly sampled from the training data and each dialogue history may contain up to $j$-$th$ turn of conversation. The actual examples of the knowledge retrieval prompt are shown in Table~\ref{table: KR}. Lastly, we retrieve the full knowledge triples using the entity and selected relation. Our knowledge retrieval agent provides a fast way to interface LLMs with external knowledge bases but is limited to one-hop reasoning due to the nature of using a single relation for knowledge retrieval.

For the item-based knowledge, which contains multiple knowledge with the same relation (e.g., \textit{[Cecilia–Star in–<movie 1, movie 2, ..., movie n>]}), we randomly select 50 knowledge triples due to the limit of input token length and only evaluate the correctness of ``Entity-Relation'' for the item-based knowledge because there is only one ground-truth knowledge for each utterance in DuRecDial dataset \cite{liu-etal-2021-durecdial2}. 

\begin{figure}[!t]
\begin{center}
\includegraphics[width=0.48\textwidth]{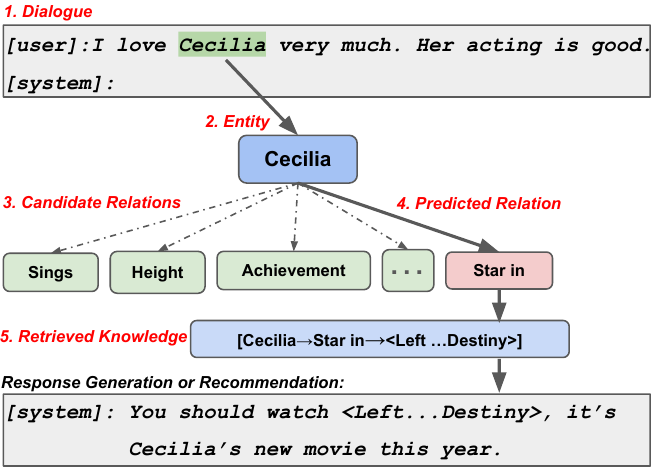}
\caption{An example of the knowledge retrieval agent.}
\label{KG-b}
\end{center}
\end{figure}

\subsection{Detailed Goal Planning Agent} \label{AG}
Both DuRecDial and TG-Redial datasets have full annotation for the goal types of each utterance. For DuRecDial, each utterance is only related to a single dialogue goal (e.g., `Asking questions' or `movie recommendation') while for TG-Redial, one utterance can have multiple dialogue goals (e.g., `Chit-chat and Asking for recommendation'), which makes it more challenging. Our goal planning uses the dialogue history to prompt the LoRA model in generating the dialogue goals for the next utterance by selecting one or multiple goals from the given goal list. The prompt is ``Given the dialogue history $C_j$, plan the next dialogue goal for the next conversation turns by selecting the dialogue goal $G$ from the <Dialogue Goal List>'' and the real example of training samples with a prompt is shown in Table~\ref{table: GP}. 
We use the full training data (around 6K and 8K for DuRecDial and TG-ReDial) in each dataset for the fine-tuning LLaMA-7b using LoRA, enhancing parameter efficiency \cite{qlora, uniCRS}. The LoRA attention dimension and scaling alpha were set to 16. While the language model was kept frozen, the LoRA layers were optimized using the AdamW. The model was fine-tuned over 5 epochs, with a batch size of 8 and a learning rate of 1 × 10-4. We compare the goal predictions results of ChatCRS with previous LM baselines like BERT \cite{BERT} and BERT+CNN \cite{uniCRS}, as well as LLM baselines like ChatGPT and LLaMA, as shown in Table~\ref{table: goal}. 

\subsection{Human Evaluation} \label{A2}
We selected 100 dialogues from the DuRecDial dataset to evaluate the performance of four methodologies: ChatGPT\footnote{\href{https://openai.com/}{OpenAI API: gpt-3.5-turbo}}, LLaMA-13b\footnote{\href{https://huggingface.co/meta-LLaMA/LLaMA-2-13b-chat-hf}{Hugging Face: LLaMA2-13b-hf}}, UniMIND, and ChatCRS. Each response generated by these methods was assessed by three annotators using a scoring system of {0: bad, 1: ok, 2: good} across four metrics: Fluency ($F_h$), Coherence ($C_h$), Informativeness ($I_h$), and Proactivity ($P_h$). The annotators, fluent in both English and Mandarin, are well-educated research assistants. This human evaluation process received IRB exemption, and the dataset used is publicly accessible. The criteria for evaluation are as follows:
\begin{itemize}[leftmargin=*]
\setlength\itemsep{0.01cm}
\item \textbf{\textit{General Language Quality:}}
\begin{itemize}[leftmargin=*]
    \item \textbf{\textit{Fluency:}}  It examines whether the responses are articulated in a manner that is both grammatically correct and fluent.
\item \textbf{\textit{Coherence:}} This parameter assesses the relevance and logical consistency of the generated responses within the context of the dialogue history. 
\end{itemize}
\item \textbf{\textit{CRS-specific Language Quality:}}
\begin{itemize}[leftmargin=*]
\item \textbf{\textit{Informativeness:}} This measure quantifies the depth and breadth of knowledge or information conveyed in the generated responses.
\item \textbf{\textit{Proactivity:}} It assesses how effectively the responses anticipate and address the underlying goals or requirements of the conversation.
\end{itemize}
\end{itemize}

Human evaluation results and an ablation study, detailed in Table~\ref{table: human}, show that ChatCRS delivers state-of-the-art (SOTA) language quality, benefiting significantly from the integration of external knowledge and goal-oriented guidance to enhance informativeness and proactivity.

\begingroup
\setlength{\tabcolsep}{3pt} 
\renewcommand{\arraystretch}{1.2} 
\begin{table}[t!]
\small
\centering
\begin{tabular}{lcccccccc}
\toprule
\multirow{2}{*}{\textbf{Model}}  & \multicolumn{4}{c}{\textbf{DuRecDial}}& \multicolumn{4}{c}{\textbf{TG-RecDial}} \\
\cmidrule(lr){2-5}  \cmidrule(lr){6-9} 
        {} & Acc & P  & R & F1 &Acc  & P  & R & F1  \\\midrule

MGCG &  NA&0.76&0.81&0.78& NA& 0.75 & 0.81 & 0.78\\
BERT & NA & 0.87 & 0.90 & 0.89 & NA & 0.92& 0.93& \textbf{0.92}\\
BERT+CNN & NA& 0.87&0.92&0.90& NA &\textbf{0.93}&\textbf{0.94}&\textbf{0.92}\\
UniMIND &  NA&0.89&0.94&0.91 & NA& {0.89} & \textbf{0.94} &{0.91}\\
ChatGPT & 0.31 & 0.05&0.04&0.04 & 0.38&0.14 & 0.10 & 0.10 \\
LLaMA &  0.11 &0.03&0.02&0.02 & 0.25 & 0.06 & 0.06& 0.05 \\
\textbf{ChatCRS} & \textbf{0.98}&\textbf{0.97}&\textbf{0.97}&\textbf{0.97}& \textbf{0.94} & 0.82 & 0.84 & 0.81 \\
\bottomrule
\end{tabular}
\caption{Results of goal planning task.}
\label{table: goal}
\end{table}
\endgroup
\subsection{Discussion on Goal Predictions} \label{A-goal}

\begin{table*}[!ht]
\small
\begin{tabular}{ p{15.5cm}} 
 \toprule{\textbf{$\spadesuit$ Examples of Single Prompt Design for the Knowledge Retrieval Agent}} \\\midrule
 \textbf{General Instruction:}\\ {You are an excellent knowledge retriever who helps select the relation of a knowledge triple [entity-relation-entity] from the given candidate relations. Your task is to choose only one relation from the candidate relations mostly related to the conversation and probably to be discussed in the next dialogue turn, given the entity and the dialogue history. Please directly answer the question in the following format: ``The relation is XXX.'',}\\\\
 \textbf{Dialogue History: } \newline[user]: Hello, Mr.Chen! How are you doing?\newline
 [system]: Hello! Not bad. It's just that there's a lot of pressure from study.\newline
 [user]:You should find a way to relax yourself properly, such as jogging, listening to music and so on.\newline ... \newline
 [system]:Well, I don't want to watch movies now.\newline
 [user]:It's starred by Aaron Kwok, who has won the Hong Kong Film Awards for Best Actor.\\ \\
 \textbf{Entity:}  Aaron Kwok \\\\
 \textbf{Candidate Relations:} \newline[`Intro', `Achievement', `Stars', `Awards', `Height', `Star sign', `Comments', `Birthplace', `Sings', `Birthday'] \\\\
 \textbf{Output: } ``The relation is Intro.''\\\\
 \toprule{\textbf{$\spadesuit$ Examples of 3-shot ICL prompt}} \\\midrule
  \textbf{Input: } (\textbf{Words in brackets provide explanations and are omitted in the actual ICL prompt})\\
  General Instruction: ... \textit{(general instruction for knowledge retrieval agent)}\\
  Dialogue History 1: ... \textit{(dialogue example from \textbf{training data})}\\
  Entity 1: ... \textit{(entity in the last utterance of dialogue history 1)}\\
  Candidate Relations 1: ... \textit{(candidate relations of entity 1)}\\
  Output 1: ... \textit{(the ground-truth relation prediction)}\\ \\
  General Instruction: ...\textit{(...)}\\
  Dialogue History 2: ... \textit{(dialogue example from \textbf{training data})}\\
  Entity 2: ...\textit{(...)}\\
  Candidate Relations 2: ...\textit{(...)}\\ 
  Output 2: ...\textit{(...)} \\ \\
  General Instruction: ...\textit{(...)}\\
  Dialogue History 3: ... \textit{(dialogue example from \textbf{training data})}\\
  Entity 3: ...\textit{(...)}\\
  Candidate Relations 3: ...\textit{(...)}\\ 
  Output 3: ...\textit{(...)} \\ \\
  \textbf{General Instruction:} ... \textit{(general instruction for knowledge retrieval agent)}\\
  \textbf{Dialogue History T:} ... \textit{(testing dialogue from \textbf{testing data})}\\
  \textbf{Entity T:} ... \textit{(entity in the last utterance of dialogue history T)}\\
  \textbf{Candidate Relations T:} ... \textit{(candidate relations of entity T)}\\ \\

  \textbf{Output: } ``The relation is XXX'' (the final relation prediction for testing data) \\\\

\bottomrule
\end{tabular}
\caption{Example of prompt design in Knowledge Retrieval Agent.}
\label{table: KR}
\end{table*}

\begin{table*}[!ht]
\small
\begin{tabular}{ p{15.5cm}} 
 \toprule{\textbf{$\spadesuit$ Examples of Prompt Design for Goal Planning Agent }} \\\midrule
 \textbf{General Instruction:}  "You are an excellent goal planner and your task is to predict the next goal of the conversation given the dialogue history. For each dialogue, choose one of the goals for the next dialogue utterance from the given goal list: \newline[``Ask about weather'', ``Food recommendation, ...,  ``Ask questions''].\\ \\
  
  \textbf{Dialogue history}\newline {[user]: I like Cecilia Cheung very much. Her acting is very good.}\newline {\textbf{.  . .}}\newline {[system]: Yeah, have you seen Cecilia Cheung's One Night in Mongkok?}\newline {[user]: I've seen it. I don't want to see it again.} \\ \\
 \textbf{Output: } ``The dialogue goal is Movie recommendation''. \\\\

\bottomrule
\end{tabular}
\caption{Example of prompt design in Goal Planning Agent.}
\label{table: GP}
\end{table*}

Figure~\ref{GOALS} illustrates the five primary goal categories along with their respective predictions across each dataset and Table~\ref{table: goal} shows the overall results of goal planning in different models for both datasets. ChatCRS demonstrates high proficiency in predicting overall goals, achieving accuracy rates of 98\% and 94\% for the DuRecDial and TG-Redial datasets respectively. Within the DuRecDial dataset, ChatCRS shows commendable performance in accurately predicting both non-recommendation goals (``say goodbye'' and ``chat about stars'') and recommendation-specific goals (``movie or music recommendation''), surpassing all comparative baselines. However, in the TG-Redial dataset, characterized by multiple dialogue goals within each utterance, ChatCRS exhibits a slight decline in accuracy for non-recommendation goals (general conversation) compared to recommendation-centric goals, leading to diminished overall accuracy.

\subsection{Baselines and Experiment Settings}\label{A1}

For the response generation and knowledge retrieval tasks in CRS, we consider the following baselines for comparisons: 
\begin{itemize}[leftmargin=*]
\setlength\itemsep{0.01cm}
\item \textbf{\textit{MGCG:}} Multi-type GRUs for the encoding of dialogue context, goal or topics and generation of response, focusing only on the response generation task \cite{liu_towards_2020_DuRecDial}.
\item \textbf{\textit{UNIMIND:}} Multi-task training framework for goal and topic predictions, as well as recommendation and response generation, focusing on both CRS tasks \cite{uniCRS}. 
\item \textbf{\textit{MGCG-G:}} GRU-based approach for graph-grounded goal planning and goal-guided response generation, focusing only on the response generation task \cite{baseline_MGCG}. 
\item \textbf{\textit{TPNet:}} Transformer-based dialogue encoder and graph-based dialogue planner for response generation and goal-planning, focusing only on response generation task \cite{baseline_TPNet}.

\end{itemize}

Additionally, we consider the following baselines for recommendation and goal-planning tasks: 
\begin{itemize}[leftmargin=*]
\setlength\itemsep{0.01cm}
\item \textbf{\textit{SASRec:}} Transformer-based recommendation system for item-based recommendation without conversations \cite{liu_towards_2020_DuRecDial}.
\item \textbf{\textit{BERT:}} BERT-based text-classification task for predicting the goal types given dialogue context \cite{BERT}.
\item \textbf{\textit{BERT+CNN:}} Deep learning approach that use the representation from MGCG and BERT for next goal predictions \cite{uniCRS}.
\end{itemize}

In our Empirical Analysis and Modelling Framework, we implement few-shot learning across various Large Language Models (LLMs) such as ChatGPT\footnote{\href{https://openai.com/}{OpenAI API: gpt-3.5-turbo-1106}}, LLaMA-7b\footnote{\href{https://huggingface.co/meta-LLaMA/LLaMA-2-7b-hf}{Hugging Face: LLaMA2-7b-hf}}, and LLaMA-13b\footnote{\href{https://huggingface.co/meta-LLaMA/LLaMA-2-13b-chat-hf}{Hugging Face: LLaMA2-13b-hf}} for tasks related to response generation and recommendation in Conversational Recommender Systems (CRS). This involves employing N-shot In-Context Learning (ICL) prompts, based on \citet{ICL}, where $N$ training data examples are integrated into the ICL prompts in a consistent format for each task. Specifically, for recommendations, the LLMs are prompted to produce a top-$K$ item ranking list (\S~\ref{Prompt}), focusing solely on the knowledge-guided generation because of the fixed dialogue goal of ``Recommendations'' and we also omit the ablation study of goal type for recommendation task.

\begingroup
\setlength{\tabcolsep}{5pt} 
\renewcommand{\arraystretch}{1} 
\begin{table}[!t]
\small
    \centering
    \begin{tabular}{lcccc}
        \toprule
        \multirow{2}{*}{\textbf{Dataset}} & \multicolumn{2}{c}{\textbf{Statistics}} & \multicolumn{2}{c}{\textbf{External K\&G}}\\
        \cmidrule(lr){2-3} \cmidrule(lr){4-5} 
        {} & Dialogues & Items & Knowledge  & Goal \\
        \midrule
        \textbf{\textit{DuRecDial}} & 	$10k$ & 	$11k$ & 	\ding{51}  & $21$ \\
        \textbf{\textit{TG-Redial}} & $10k$ & $33k$ & \color{red}{\ding{55}} & $8$ \\
        \bottomrule
    \end{tabular}
    \caption{Statistics of datasets}
    \label{data}
\end{table}
\endgroup

For the Modelling Framework's goal planning agent, QLora is utilized to fine-tune LLaMA-7b, enhancing parameter efficiency \cite{qlora, uniCRS}. The LoRA attention dimension and scaling alpha were set to 16. While the language model was kept frozen, the LoRA layers were optimized using the AdamW. The model was fine-tuned over 5 epochs, with a batch size of 8 and a learning rate of 1 × 10-4. 
The knowledge retrieval agent and LLM-based generation unit employ the same N-shot ICL approach as in CRS tasks with ChatGPT and LLaMA-13b \cite{structgpt}. Given that TG-Redial \cite{zhou_towards_2020_TGRedial} comprises only Chinese conversations, a pre-trained Chinese LLaMA model is used for inference\footnote{\href{https://huggingface.co/seeledu/Chinese-LLaMA-2-7B}{Hugging Face: Chinese-LLaMA2}}. Our experiments, inclusive of LLaMA, UniMIND or ChatGPT, run on a single A100 GPU or via the OpenAI API. The one-time ICL inference duration on DuRecDial \cite{liu-etal-2021-durecdial2} test data spans 5.5 to 13 hours for LLaMA and ChatGPT, respectively, with the OpenAI API inference cost approximating US\$20 for the same dataset. Statistics of two experimented datasets are shown in Table~\ref{data}.

\begin{table*}[ht!]
\small
\begin{tabular}{ p{15.5cm}} 
 \toprule{\textbf{$\spadesuit$ Examples of Prompt Design for Empirical Analysis }} \\\midrule
 \textbf{General Instruction:} {You are an excellent conversational recommender who helps the user achieve recommendation-related goals through conversations.}\\\midrule
  \textbf{DG Instruction on Response Generation Task:} You are an excellent conversational recommender who helps the user achieve recommendation-related goals through conversations. Given the dialogue history, your task is to generate an appropriate system response. Please reply by completing the output template ``The system response is []''\\\midrule
  \textbf{DG Instruction on Recommendation Task:} 
  You are an excellent conversational recommender who helps the user achieve recommendation-related goals through conversations. Given the dialogue history, your task is to generate appropriate item recommendations. Please reply by completing the output template ``The recommendation list is [].'' Please limit your recommendation to 50 items in a ranking list without any sentences. If you don't know the answer, simply output [] without any explanation.\\\midrule
  
\textbf{COT-G Instruction on Response Generation Task:} You are an excellent conversational recommender who helps the user achieve recommendation-related goals through conversations. Given the dialogue history, your task is to first plan the next goal of the conversation from the goal list and then generate an appropriate system response. Goal List: [ ``Ask about weather'', ``Food recommendation'', ``POI recommendation'', ... ,  ``Say goodbye'']. Please reply by completing the output template ``The predicted dialogue goal is [] and the system response is []''.\\\midrule

\textbf{COT-K Instruction on Response Generation Task:} You are an excellent conversational recommender who helps the user achieve recommendation-related goals through conversations. Given the dialogue history, your task is to first generate an appropriate knowledge triple and then generate an appropriate system response. If the dialogue doesn't contain knowledge, you can directly output ``None''. Please reply by completing the output template ``The predicted knowledge triples is [] and the system response is [].''\\\midrule
\textbf{COT-K Instruction on Recommendation Task:} You are an excellent conversational recommender who helps the user achieve recommendation-related goals through conversations. Given the dialogue history, your task is to first generate an appropriate knowledge triple and then generate appropriate item Recommendations. If the dialogue doesn't contain knowledge, you can directly output ``None''. Please reply by completing the output template ``The predicted knowledge triples is [] and the recommendation list is []''. Please limit your recommendation to 50 items in a ranking list without any sentences. If you don't know the answer, simply output [] without any explanation.\\\midrule

\textbf{Oracle-G Instruction on Response Generation Task:} You are an excellent conversational recommender who helps the user achieve recommendation-related goals through conversations. Given the dialogue history and the dialogue goal of the next system response, your task is to first repeat the conversation goal and then generate an appropriate system response. Please reply by completing the output template ``The predicted dialogue goal is [] and the system response is []''.\\\midrule

\textbf{Oracle-K Instruction on Response Generation Task:} You are an excellent conversational recommender who helps the user achieve recommendation-related goals through conversations. Given the dialogue history and knowledge triple for the next system response, your task is to first repeat the knowledge triple and then generate an appropriate system response. Please reply by completing the output template ``The predicted knowledge triples is [] and the system response is [].''\\\midrule

\textbf{Oracle-K Instruction on Recommendation Task:} You are an excellent conversational recommender who helps the user achieve recommendation-related goals through conversations. Given the dialogue history and knowledge triple for the next system response, your task is to first repeat the knowledge triple and then generate appropriate item Recommendations. Please reply by completing the output template ``The predicted knowledge triples is [] and the recommendation list is []''. Please limit your recommendation to 50 items in a ranking list without any sentences. If you don't know the answer, simply output [] without any explanation.\\\midrule
\textbf{Oracle-BOTH Instruction on Response Generation Task:}  You are an excellent conversational recommender who helps the user achieve recommendation-related goals through conversations. Given the dialogue history, the conversation goal and knowledge triple for the next system response, your task is to first repeat the conversation goal and knowledge, and then generate appropriate item Recommendations. Please reply by completing the output template ``The predicted dialogue goal is [], the predicted knowledge is [] and the system response is []''. \\
\bottomrule
\end{tabular}
\caption{Example of instruction in prompt design for Empirical Analysis.}
\label{p:ins}
\end{table*}

\begin{table*}[t!]
\small
\begin{tabular}{ p{15.5cm}} 

 \toprule{\textbf{$\spadesuit$ Examples of Input and Output Format in Prompt Design for Empirical Analysis}} \\\midrule

     \textbf{Dialogue History:} \newline {[user]:Hello! Do you know who starred in the movie Flying Dagger?\newline [system]: Yes, of course I know that. It's Jimmy Lin.\newline [user]: OK, thank you.\newline [system]: He is an amazing all-rounder, and he won Chinese Youth Leader in 2014.\newline [user]: He is my favourite star.\newline [system]: }\\ \\
    \textbf{Dialogue Goal:} Movie recommendation \\ \\
    \textbf{Knowledge:} `Jimmy Lin', `Stars', `To Miss with Love' \\ \\
    \textbf{Ground-Truth Recommendation:} To Miss with Love \\ \\
    \textbf{Ground-Truth Response:} Since you like him so much, I wanna recommend to you the movie To Miss with Love, which is starred by him.\\
    
    \midrule

  \textbf{DG Input and Output on Response Generation Task:} \\Input: Dialogue History; \\Output: ``The system response is []''\\\midrule
  \textbf{DG Input and Output on Recommendation Task:} \\Input: Dialogue History; \\Output: ``The recommendation list is [].'' \\\midrule
  
\textbf{COT-G Input and Output on Response Generation Task:} \\Input: Dialogue History; \\Output: ``The predicted dialogue goal is [] and the system response is []''.\\\midrule

\textbf{COT-K Input and Output on Response Generation Task:}\\ Input: Dialogue History; \\Output: ``The predicted knowledge triple is [] and the system response is [].''\\\midrule
\textbf{COT-K Input and Output on Recommendation Task:}\\ Input: Dialogue History; \\Output: ``The predicted knowledge triple is [] and the recommendation list is []''.\\\midrule

\textbf{Oracle-G Input and Output on Response Generation Task:}\\ Input: Dialogue History + Dialogue Goal; \\Output: ``The predicted dialogue goal is [] and the system response is []''.\\\midrule

\textbf{Oracle-K Input and Output on Response Generation Task:} \\ Input: Dialogue History + Knowledge; \\Output: ``The predicted knowledge triple is [] and the system response is [].''\\\midrule

\textbf{Oracle-K Input and Output on Recommendation Task:} \\ Input: Dialogue History + Knowledge; \\Output: ``The predicted knowledge triple is [] and the recommendation list is []''. \\\midrule
\textbf{Oracle-BOTH Input and Output on Response Generation Task:}  \\ Input: Dialogue History + Dialogue Goal + Knowledge; \\Output: ``The predicted dialogue goal is [], the predicted knowledge is [] and the system response is []''. \\
   
\bottomrule
\end{tabular}
\caption{Example of input and output format in prompt design for Empirical Analysis.}
\label{p:out}
\end{table*}

\end{document}